\documentclass[10pt,twocolumn,letterpaper]{article}

\usepackage[pagenumbers]{arxiv}

\usepackage{graphicx}
\usepackage{amsmath}
\usepackage{amssymb}
\usepackage{times}
\usepackage{epsfig}
\usepackage{booktabs}
\usepackage{mathtools}  % for '\DeclareMathOperator' macro
\DeclareMathOperator{\Tr}{Tr}
\usepackage{dirtytalk}
\usepackage{caption}
\newcommand\norm[1]{\left\lVert#1\right\rVert}
\usepackage[acronym,toc,shortcuts]{glossaries}
\usepackage{multirow}
\usepackage{microtype}
\usepackage{url}
\usepackage{xcolor}

\newacronym{dl}{DL}{Deep Learning}
\newacronym{dp}{DP}{Differential Privacy}
\newacronym{dpsgd}{DP-SGD}{Differentially Private Stochastic Gradient Descent}
\newacronym{cnn}{CNNs}{Convolutional Neural Networks}
\newacronym{ecnn}{ECNN}{Equivariant CNN}
\newacronym{wrn}{WRN}{WideResNet}
\newacronym{fir}{FIR}{Filter Impulse Response}
\newacronym{ema}{EMA}{Exponential Moving Average}
\newacronym{sota}{SOTA}{state-of-the-art}

\usepackage[capitalize]{cleveref}
\crefname{section}{Sec.}{Secs.}
\Crefname{section}{Section}{Sections}
\Crefname{table}{Table}{Tables}
\crefname{table}{Tab.}{Tabs.}

\begin{document}

%%%%%%%%% TITLE
\title{Equivariant Differentially Private Deep Learning:\\Why DP-SGD Needs Sparser Models}

\author{Florian A. Hölzl, Daniel Rueckert, Georgios Kaissis\\
Institute for Artifical Intelligence in Medicine, Technical University of Munich\\
{\tt\small \{florian.hoelzl, daniel.rueckert, g.kaissis\}@tum.de}}

\maketitle

%%%%%%%%% ABSTRACT
\begin{abstract}
    \glsfirst{dpsgd} limits the amount of private information deep learning models can memorize during training.
    This is achieved by clipping and adding noise to the model's gradients, and thus networks with more parameters require proportionally stronger perturbation.
    As a result, large models have difficulties learning useful information, rendering training with \gls{dpsgd} exceedingly difficult on more challenging training tasks.
    Recent research has focused on combating this challenge through training adaptations such as heavy data augmentation and large batch sizes.
    However, these techniques further increase the computational overhead of \gls{dpsgd} and reduce its practical applicability.
    In this work, we propose using the principle of sparse model design to solve precisely such complex tasks with fewer parameters, higher accuracy, and in less time, thus serving as a promising direction for \gls{dpsgd}.
    We achieve such \textit{sparsity by design} by introducing equivariant convolutional networks for model training with \glsfirst{dp}.
    Using equivariant networks, we show that small and efficient architecture design can outperform current \gls{sota} with substantially lower computational requirements.
    On CIFAR-10, we achieve an increase of up to $9\%$ in accuracy while reducing the computation time by more than $85\%$.
    Our results are a step towards efficient model architectures that make optimal use of their parameters and bridge the privacy-utility gap between private and non-private deep learning for computer vision.
\end{abstract}

%%%%%%%%% BODY TEXT
\begin{figure*}[!t]
\includegraphics[width=\textwidth]{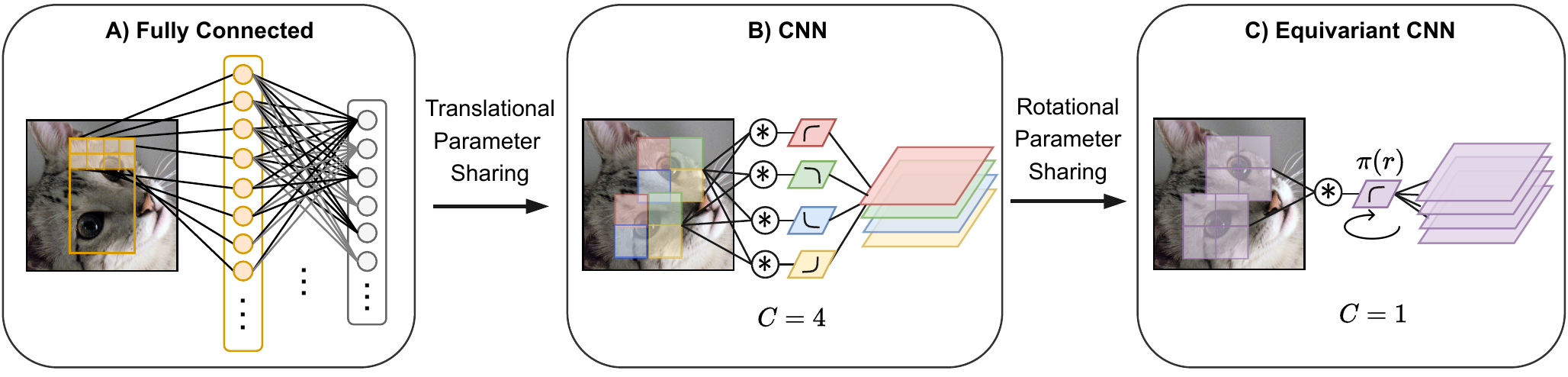}
\caption{
Sparsity by design indicates that fewer overall parameters are required to learn the same representations from an input image. 
Conventional \gls{cnn} achieve this by applying convolutions with small kernels across the image instead of flattening and densely connecting all pixels.
Equivariant convolutions introduce additional transformations on the kernel, allowing for the detection of features independent of their \eg rotation and/or reflection (pose).
As a result, equivariant networks need even fewer parameters to learn the same information and are thus even sparser by design.
Image taken from the ImageNet dataset.
}\label{fig:tmlr_sparsity_figure}
\end{figure*}

\section{Introduction and Related Work}\label{sec:intro}
Artificial Intelligence is increasingly applied to fields where extremely sensitive data is used, such as medicine or the social sciences. 
Previous research has demonstrated that sensitive information can be reverse-engineered from unprotected machine learning models~\cite{balle2022, carlini2021}.
This renders privacy concerns a major hurdle to developing and deploying machine learning systems in fields where the protection of sensitive data is necessary, \eg due to privacy regulations, intellectual property requirements or other ethical considerations and necessitates steps preventing the exposure of such data to unauthorized third parties.

Privacy-enhancing technologies allow one to derive insights from sensitive datasets while quantitatively bounding the risk of information leakage about the training samples. 
By giving formal guarantees on data protection, they represent the best chance to date to incentivize data sharing in an ethical and responsible manner.
\gls{dp}~\cite{dwork2014}, the gold-standard technique for privacy preservation, is most commonly applied in the field of deep learning by utilizing \gls{dpsgd}~\cite{abadi2016}, which imposes an upper bound on how much information can be extracted from the model's weights about the individuals whose data was used for training.
This guarantee is achieved during model training through a modified optimization algorithm, which includes clipping the gradient vector based on a predefined threshold of its $\ell_2$-norm and then adding noise to it (\textit{see \cref{sec:background}}).
Such a gradient is then considered \textit{privatized}.
A predictive model trained with privatized gradients is subsequently private as well, allowing analysts to share the model while retaining the ability to bound the capacity of adversaries to derive information from the data used for training.

However, using gradients privatized with \gls{dpsgd} often leads to sharp reduction in prediction performance.
This problem is called the \textit{privacy-utility trade-off} and is due to two main factors:
On one hand, clipping the gradient diminishes its information content and biases its direction; moreover, the total noise magnitude scales proportionally to the $\ell_2$-norm of the gradient, and thus to the number of model parameters.
This leads to a decrease in the signal-to-noise ratio, in that the true gradient's signal is diminished relative to the noise introduced by \gls{dpsgd}.
Consequently, the effective training of large models, typically used to achieve \gls{sota} results from scratch in non-private training, is rendered disproportionately difficult and falls far short of attaining comparable results to non-private training.
In this paper, we propose to address the privacy-utility trade-off by introducing \textit{sparse model design through \gls{ecnn}} architectures.

\subsection{Computational Considerations of DP-SGD}
Recent advancements in \gls{dp} deep learning have focused largely on optimizing the training regime of larger models to mitigate the previously mentioned difficulties.
The current \gls{sota} for training CIFAR-10 from scratch was recently established by \cite{de2022} and incrementally improved by~\cite{sander2022}. 
These works' techniques focus on extensive training adaptions (which likely have an effect on the smoothness of the optimisation landscape) to improve prediction performance.
Among the most important adaptations are (1) large-batch training~\cite{doermann2021}, (2) averaging per-sample gradients across multiple augmentation of the same image before clipping (\textit{augmentation multiplicity})~\cite{hoffer2020, fort2021}, and (3) temporal parameter averaging techniques~\cite{polyak1992}.
This results in remarkable accuracy gains on over-parameterized models over the previous \gls{sota}, which seemed out of reach up until recently for models such as the \gls{wrn}, that incorporate several million parameters.
However, the accuracy gains presented by the authors of the aforementioned work come at a high cost: the massive computational resources and time required to train those models have an overbearing financial (and environmental) impact and impose great difficulties in reproducing the results. 
As a result, further progress in this direction is cumbersome and inefficient, and renders proposing improvements on top of the presented results out of reach, especially for scientific institutions without access to large-scale computational resources.
In summary, current \gls{sota} models trained with \gls{dpsgd} (1) in part still substantially lack behind their non-private counterparts and (2) have an impracticable computational burden that makes research improvements difficult.

Orthogonal to the aforementioned works, other studies in \gls{dp} deep learning have introduced training regimes which additionally leverage publicly available data, which is ostensibly usable without any privacy constraints~\cite{golatkar2022, yu2021}.
By starting from a well performing non-private base model, fine-tuning on private datasets with \gls{dpsgd} can lead to promising results close to non-private training from scratch.
However, \textit{Tramèr} \etal question if using public data for pre-training should actually be considered differential-privacy-preserving as pre-trained models can leak private information contained within the public dataset and thus negatively impact the public perception of the field~\cite{tramer2022}.
Relying on public data in many cases thus opposes the very foundation of privacy-enhancing technologies, \ie obtaining access to important \textit{insights} while minimizing access to the sensitive data required.  
Even more importantly for practical application, relying on pre-training as a solution is not a panacea.
Large quantities of public data are not available in areas such as medicine, where data, at least from the same distribution, still comes with the same privacy considerations and (even when available in sufficient quantity) is difficult to access.
Not only has the importance of in-distribution data been shown in previous works~\cite{he2019, radimagenet}, but relying on pre-training alone increases the barrier to extend \gls{dpsgd} to new modalities and novel prediction tasks.
Pre-training and its positive impact showcased on benchmark datasets is thus an interesting observation, but will not solve the previously mentioned challenges that currently hold back \gls{dpsgd} from widespread adaptation in practice.

\subsection{Motivation}
We thus contend that an optimal solution to the aforementioned dilemmas will require a fundamental reconsideration, which will marry \textit{high prediction performance} when training from scratch with \textit{high computational efficiency}.
The guiding hypothesis of our work is, that networks that exhibit designed sparsity can achieve the aforementioned goal.
As we demonstrate in this work, the notion of \textbf{designed sparsity}, introduced by \cite{hoefler2021}, combines two major characteristics beneficial for \gls{dpsgd}: (1) increased representational efficiency and (2) a reduced set of (possibly redundant) model parameters.  
The requirement for \gls{dp} models to learn high-quality features to achieve parity with non-private models has been previously discussed~\cite{tramer2020}.
However, the aforementioned work utilises a cascade of static feature extractors, which lack the flexibility of their learned counterparts and whose capability to scale to more complex problems remains an open research question.
In contrast, we show that \textbf{networks that are sparse by design remain trainable with a similar or higher degree of expressiveness, but with fewer parameters} than comparable deep learning models.
In fact, even conventional \gls{cnn} are an example of such networks, since a single kernel is used to compute the features of a whole image.
They inherently exhibit parameter sharing with respect to translations and can thus be seen as a \textit{sparse version} of a fully connected layer.
We argue that, through additional improvements in model architectures that lead to higher designed sparsity, networks can learn features more efficiently, offering a promising direction for \gls{dpsgd}.
To evaluate this hypothesis, we introduce \gls{ecnn}s for \gls{dpsgd}.
As shown in \cref{fig:tmlr_sparsity_figure}, \gls{ecnn}s further increase parameter sharing by being equivariant to transformations such as rotations and reflections.
They thus offer an even higher degree of designed sparsity and a possible solution for efficient training under \gls{dp}.

While rotational equivariance can be approximated \textit{with no formal guarantee} by conventional (non-equivariant) \gls{cnn} through an increase in model width, dataset size and augmentation techniques, \ie the exact techniques mentioned earlier, this approximation has two important drawbacks.
They (1) massively increase the computation time of the (already very demanding) \gls{dpsgd}, as \eg each additional set of simultaneous augmentations increases the time complexity almost linearly. 
And (2), na\"ively scaling the number of parameters proportionally increases the total noise power of the added Gaussian noise, thus risking \say{drowning out} the learning signal.
Network layers equivariant to rotation and reflection transformations preserve the relative \textit{pose} of features in addition to the translational equivariance preserved by standard convolutions.
The resulting additional parameter sharing thus avoids the redundant learning of identical convolutional filters for multiple poses.
In addition to their high parameter efficiency, \gls{ecnn} architectures are known to exhibit increased data efficiency and improved generalization, especially in domains with high degrees of intra-image symmetry~\cite{cohen2016, cohen2017}.
So far, however, no works have investigated how to combine equivariant layers with \gls{dp} training nor analyzed the potential beneficial changes to the training regime, even though their characteristics render them highly attractive for this use-case.

In this work, we show the \textbf{need for sparse model architectures under \gls{dpsgd}}, and that \gls{ecnn}s are a promising step in this direction.
Their higher designed sparsity compared to standard \gls{cnn}, allows them to outperform larger models while simultaneously decreasing the required computation time.
As we will demonstrate, this renders them especially interesting for training with \gls{dpsgd}, particularly under tighter privacy bounds and low data regimes, both desirable traits for privacy-preserving techniques.

Our main contributions are summarized as follows:
\begin{itemize} 
    \item We introduce the methodology necessary to train \gls{ecnn}s with \gls{dpsgd} in \cref{sec:method}.
    As part of this contribution, we propose novel normalization layers for discrete and continuous symmetry groups that preserve the equivariance property \textit{and} fulfill the \gls{dp} condition.
    \item By leveraging the notion of designed sparsity, we substantially improve the current \gls{sota} on \gls{dpsgd} imaging benchmarks without additional data in \cref{sec:results}.
    We experimentally demonstrate \gls{ecnn}s as a promising architecture that satisfies this concept by offering improved results while requiring fewer parameters than a conventional \gls{cnn}.
    Among others, we show an increase of $9.2\%$ under $(2,10^{-5})$-\gls{dp} on CIFAR-10 and an increase of more than $10\%$ on Tiny-ImageNet-200.
    \item We provide insights into model calibration of our approach, since poor calibration is a known weakness of \gls{dpsgd}.
    We find that the proposed equivariant architecture improves model calibration with an on average $17\%$ lower Brier score across all evaluated datasets compared to the conventional network.
    \item We experimentally show, that equivariant networks are more robust to key hyperparameter choices from recent \gls{sota} results, in particular augmentations in the input domain and batch size.
    Additionally, we analyze how equivariant specific hyperparameter choices, such as the symmetry groups, affect training with \gls{dpsgd} in \cref{sec:ablation}.
\end{itemize}

%-------------------------------------------------------------------------
\section{Background}
\label{sec:background}
\subsection{Differential Privacy}
At its core, differential privacy provides a way to answer questions about a dataset while limiting the risk of revealing sensitive information about specific samples in the dataset.
It achieves this by introducing controlled randomness into the data analysis process.
In other words, \gls{dp} is a strong stability condition on randomised algorithms mandating that outputs are approximately invariant under inclusion or exclusion of a single individual from the input database.
For a mechanism (randomized algorithm) $\mathcal{M}$ and all datasets $D$ and $D^\prime$ that differ in one element as well as all measurable subsets $S$ of the range of $\mathcal{M}$, $(\varepsilon,\delta)$-\gls{dp} requires that:
\begin{equation}\label{eq:dp}
\text{Pr}(\mathcal{M}(D) \in S) \leq e^\varepsilon \, \text{Pr}(\mathcal{M}(D^\prime) \in S) + \delta,
\end{equation}
where the privacy guarantees of the algorithm are parameterised by $\varepsilon \geq 0$ and $\delta \in [0,1)$.
This privacy constraint is in practice typically realised by the addition of noise.
For a comprehensive overview of \gls{dp}, we refer to \cite{dwork2014}.
Its application to deep learning came with the introduction of \gls{dpsgd} by \cite{abadi2016}.
In our work, we utilise Gaussian noise and the aforementioned \gls{dpsgd} algorithm to privatise gradient updates in neural network training.
We use Rényi-\gls{dp} accounting~\cite{mironov2017, mironov2019} to track the privacy loss throughout the training.
Rényi-\gls{dp} (RDP) is often used with training deep neural networks as it massively facilitates the composition of sequences of private algorithms executed on sub-samples of a larger dataset, such as in SGD, where we iteratively update the model parameters using randomly selected subsamples of the training data.
The RDP privacy condition is:
\begin{equation}\label{eq:renyi_dp}
D_{\alpha}(\mathcal{M}(D) \parallel \mathcal{M}(D^\prime)) \leq \rho,
\end{equation}
where $D_{\alpha}$ is the Rényi divergence of order $\alpha \geq 1$.
We note that the Rényi divergence is symmetric in the Gaussian noise setting, as the \gls{dp}-guarantee is required to be symmetric. 
RDP can be converted to $(\varepsilon, \delta)$-DP for a given $\delta$.
In the rest of this paper, we will refer to the converted $\bar{\varepsilon}$ simply as $\varepsilon$.
We note that we refer to \say{sampling} rather than \say{mini-batches} in \gls{dpsgd}, since privacy amplification by sampling requires subsets of the training set to be drawn using \eg a Poisson sampling technique~\cite{abadi2016}.

\subsection{Equivariant CNNs}
Equivariance describes the mathematical property of a structure-preserving mapping.
This means, that there exist two transformations $T_g$ and $T^{\prime}_{g}$ that lead to the same result when applying a mapping $\phi$ to an input $\mathbf{x}$ such that:
\begin{equation}\label{eq:equivariance_prop}
\phi\left(T_g \mathbf{x}\right) = T^{\prime}_g \phi\left(\mathbf{x}\right) \ .
\end{equation}
For image classification, this formal guarantee was first introduced for rotations and reflections in Regular Group CNNs by \cite{cohen2016} and later extended to \gls{ecnn}s with steerable filters~\cite{cohen2017, cohen2018, weiler2018}.
The general approach to equivariant convolutions is centered around the idea of representations, describing the transformation laws of a given feature space.
The feature fields in this space are mappings that transform according to the corresponding representation.
Each layer's input and output space must be compatible with the corresponding transformation law.
In order to guarantee this behavior, a convolution kernel $K$ is subject to a linear constraint, given by
\begin{equation}\label{eq:equivariance_constraint}
K(g\cdot \mathbf{x}) = \rho_{\text{out}}(g) K(\mathbf{x}) \rho_{\text{in}}(g)^{-1} \; \; \forall g \in G, \; x \in \mathbb{R}^n,
\end{equation}
with group actions $g \in G$, depending on the associated group representations $\rho$.
For our case (planar images), we focused on representations where the group actions are rotations and reflections acting on the parameters of a learned kernel, \ie the group elements.
These group actions can be discrete, with the number of rotations typically denoted as $N$, or continuous in $SO(2)$ or $O(2)$.
Finite groups are commonly represented through regular representations, where the corresponding transformation matrix has dimensionality equal to the order of the group, \eg $\mathbb{R}^N$.
There are different ways to approximate continuous groups to work with steerable convolutions.
We follow the results of \cite{weiler2019}, where the $SO(2)$ group has been shown to give the most promising results of all continuous groups for non-private planar image classification.
The performance of the $SO(2)$ group is evaluated using irreducible representations.
An irreducible representation $\psi$ of a group is a representation that cannot be reduced or decomposed into smaller independent representations.
Formally, for our group $G$ and the feature space $Y$, an irreducible representation of $G$ on $Y$ is a linear map $\psi: G \mapsto GL(Y)$ to the general linear group $GL$ that preserves the group structure.
In our case, the kernel $K$ is a linear map, with its basis given by breaking down a group representation into a direct sum of irreducible representations.
This is called a irreducible representation decomposition and allows us to construct a equivariant map between pairs of irreducible representations, \ie $\rho_{\text{in}}$ and $\rho_{\text{out}}$.
To solve the corresponding kernel constraint in our equivariant convolutions, we employ the general solution from \cite{cesa2022}.

%-------------------------------------------------------------------------
\section{Methodology}
\label{sec:method}
In this section, we describe our method and introduce novel layers to allow training \gls{ecnn}s with \gls{dpsgd}.
The proposed layers preserve the notion of orientation in our \gls{ecnn}s without violating the \gls{dp} constraint.
Maintaining the equivariance property, not only within a single layer but throughout the whole network, is necessary to fully leverage the additional feature information during training.
Our approach is based on the equivariant frameworks established by~\cite{geiger2022, cesa2022}.

\subsection{Convolution Layers}
To satisfy the kernel constraint~\cref{eq:equivariance_constraint}, we define the transformation law of each convolution by its input and output representations $\rho_{in}$ and $\rho_{out}$.
For the roto-translational \gls{ecnn}s proposed in this work, the representations correspond to rotation matrices of a specific symmetry group (\ie $C_N$, $D_N$ \textit{or} $SO(2)$).
The resulting transformation law of the feature space is a constant linear mapping that only has to be computed once during initialization.
While lower-level features intrinsically exhibit a higher degree of symmetry, natural images often have a sense of orientation at a global level.
To account for this varying level of equivariance, we choose an initial symmetry group and restrict $\rho_{in}$ and $\rho_{out}$ for the last residual block.
The order of the finite groups is reduced to $N/2$, with $N$ being the group order, for the last residual block.
When choosing $SO(2)$ as the initial symmetry group, we restrict the representations in the last residual block to be invariant.
Additionally, we adjust the number of channels for each convolutional layer, such that our equivariant networks have a similar number of parameters as their non-equivariant counterparts.
This is done by multiplying the number of channels by $\sqrt{1.5 N}/N$ to end up with a similar number of parameters for all symmetry groups.
Moreover, to improve signal propagation, we apply weight standardization in the convolutional layer~\cite{qiao2019} and switch the order of the normalization and activation layers as proposed in~\cite{sander2022}.

\subsection{Normalization and Nonlinearities}
For \gls{dp}, we are required to compute per-sample gradients, which are incompatible with the batch normalization typically used in \gls{ecnn} literature.
We thus propose \textbf{two novel equivariant normalization layers} for \gls{dp}-\gls{ecnn} training.
Firstly, for trivial and regular representations used for discrete groups $C_N$ and $D_N$, we introduce the (na\"ive) \textit{equivariant group normalisation} layer.
The additional pose information is maintained throughout the network by increasing the channel dimension $C$ of the feature vector to include the fields of each representation.
Before normalization, we split the channel dimension $C$ of our $4$-dimensional feature vector such that the representation fields of each channel are in a separate dimension.
The resulting feature vector $(N,Y,C,W,H)$, with the spatial dimensions $H$, $W$ and the batch axis $N$, is used for normalization.
Each feature field $\mathbf{y}\in [1,Y]$, corresponding to a representation's transformation, is then normalized across a group of channels $C$.
This implementation allows us to utilize existing $3$-dimensional group normalisation layers. After the group normalization, for consistency, we reduce our feature vector back to $4$ dimensions by stacking the feature fields of each channel.

The aforementioned layer is, however, \textit{unsuitable for use with continuous groups}, which require irreducible representations to function.
To thus satisfy the equivariance property for the irreducible representations of the $SO(2)$ group, we introduce an \textit{i.i.d. instance normalisation} layer.
To normalize the feature fields, we require an estimate of their expected values and variance when transforming them by a group action.
In the following, we derive how to calculate the two values.
For the former, we compute the expectation over the whole group $G$ that is acting on our features.
This expectation can be written by using a normalized Haar measure $\mu$ over $G$,
\begin{equation}
    \mathbb{E}_{g\in G}[\psi(g)]=\int_G \text{d}\mu(g)\psi(g)
\end{equation}
with the irreducible representation $\psi$ of $G$~\cite{chirikjian2000}.
Due to the orthogonality of $\psi(g)\ \forall g\in G$, the integral is always zero for all non-trivial irreducible representations.
As a consequence, the mean of any vector transforming according to those representations is also zero.
It follows that the expected value of $\rho(g)$ is a null matrix with non-zero values in the diagonal if and only if $\psi_i$ is a trivial representation.
In practice, we can thus pre-compute $\mathbb{E}_{g\in G}[\psi(g)]$ when initializing our equivariant layers to calculate the estimated mean based on the trivial representations.
During training, we then simply subtract it from our calculated feature fields.

To derive the expected variance, we substitute the transformation in \cref{eq:equivariance_prop} with our representation $\rho(g)$ to receive the feature field $\mathbf{y}=\rho(g)\phi(\mathbf{x})$ of an input $\mathbf{x}$.
The covariance of $\mathbf{y}$ can then be calculated with
\begin{equation}
    \mathbb{E}_{\mathbf{y}\in Y}\left[\mathbf{yy}^{\text{T}}\right]=\mathbb{E}_{g\in G}\left[\rho(g)\, \phi(\mathbf{x})\phi(\mathbf{x})^{\text{T}}\rho(g)^{\text{T}}\right] .
    % \mathbb{E}_{\mathbf{y}\in Y}\left[\mathbf{yy}^{\text{T}}\right]=\mathbb{E}_{g\in G}\left[\rho(g)\, \mathbb{E}_{\mathbf{x}\in X}\left[\mathbf{xx}^{\text{T}}\right]\rho(g)^{\text{T}}\right] .
\end{equation}
Due to the fact that our representations $\rho(g)$ are orthonormal and irreducible, the covariance matrix must be an orthogonal matrix.
In addition, the covariance matrix is symmetric and semi-positive definite, making it a multiple of the identity matrix, \ie $\mathbb{E}_{\mathbf{y}\in Y}[\mathbf{yy}^{\top}]=\lambda \mathbf{I}$.
We can then compute this multiple independent of our representation simply as
\begin{equation}
    d\lambda=\Tr(\mathbb{E}_{\mathbf{y}\in Y}\left[\mathbf{yy}^{\text{T}}\right])=\mathbb{E}_{\mathbf{y}\in Y}\left[\norm{\mathbf{y}}_2^2\right] .
\end{equation}
Finally, we also apply a learnable weight and bias parameter as with most normalization layers.

For trivial and regular representations, standard activation functions, such as Mish and ReLU~\cite{misra2020, agarap2018}, can be used as non-linearities.
When working with the $SO(2)$ group, we use an Inverse Fourier Transform on the feature space to then apply the non-linearity in the group domain.
Afterwards, we compute the Fourier Transform to again obtain the coefficients of the irreducible representations.

\section{Results}
\subsection{Equivariant Models are Sparse by Design}\label{sec:eval_sparsity}
\subsubsection{Fewer Model Parameters}
\begin{figure}[t]
    \includegraphics[width=0.49\textwidth]{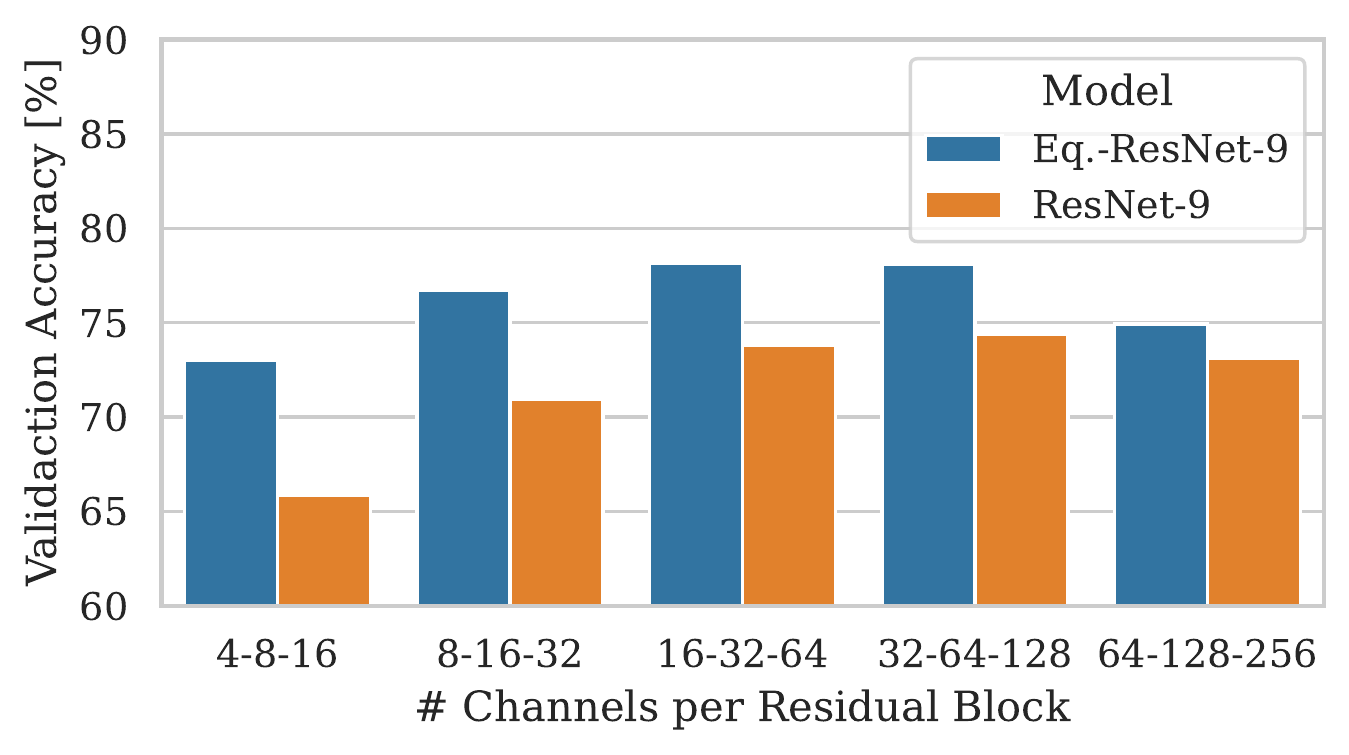}
\caption{In comparison to the non-equivariant ResNet-9, our Equivariant-ResNet-9 has a \textit{sweet spot} on CIFAR-10 at an intermediate model size of $16$, $32$ and $64$ channels per layer for the three residual blocks under $(8,10^{-5})$-\gls{dp}.}\label{fig:model_width}
\end{figure}

In previous work by \textit{Klause} \etal~\cite{klause2022}, the ResNet-9 architecture was shown to be performant with \gls{dpsgd}, despite its small size.
We are interested on how this efficiency can be further increased and thus use it as our baseline.
We find that the equivariance property of the Equivariant-ResNet-9 allows us to further increase the validation accuracy on CIFAR-10 by reducing the number of convolution channels per layer.
In fact, \cref{fig:model_width} indicates that there is an \say{optimal} model layout of $(16, 32, 64)$ channels, denoting the filters per layer in the three residual blocks.
For larger layouts, the performance starts \textit{decreasing} again.
This leads to a total number of optimal parameters of $\approx250k$, \ie \textit{ten times fewer than the original} (conventional) CNN.
In comparison, the validation accuracy of the standard ResNet-9 does not substantially increase when the model size is reduced.
This is likely due to the fewer parameters not being able to learn features effectively.
The equivariant network, on the other hand, does not have to learn redundant features for different orientations in separate channels due to the additional pose information in our feature space.
The experiments indicate that this allows the network to still detect a sufficient amount of information in the image for prediction, even when reducing the parameters in the network.
How much of this additional information can be learned per parameter, also called \textit{parameter utilization}~\cite{cohen2017}, is dependent on the chosen symmetry group.  
While the equivariance property is better satisfied for the symmetry groups $SO(2)$ and $O(2)$, the expressiveness of the trained kernel can suffer due to the corresponding tighter kernel constraints as introduced in~\cref{sec:background}.
Our ablation studies in \cref{sec:ablation} show that the dihedral $D_4$ group offers a \say{sweet spot} regarding performance and accuracy, and is thus used throughout this section.
To summarize, we tested and found that the equivariance property allows us to reduce the width of our network while maintaining or even increasing the classification accuracy with \gls{dpsgd} by increasing the parameter utilization in the network.

\subsubsection{More Parameters Contribute to Model Prediction}
In addition to an overall smaller model size, we also evaluate how the sparse design of the architecture affects the gradients and parameters, and their impact on training and prediction correspondingly.
Due to the increased parameter sharing, we expect the smaller model to use more of its parameters to make predictions.
To analyze this, we utilize the $\ell_{0}^{\epsilon}$ \say{norm}, \ie the number of parameters or gradient entries with a magnitude smaller than $\epsilon$~\cite{donoho2003}.
We note that this is a different $\epsilon$ than the one used to describe \gls{dp} guarantees.
\cref{fig:eval_sparsity} shows, that the percentage of parameters with absolute values $\approx 0$ during training is substantially lower for the equivariant network compared to the standard ResNet-9.
Thus, the equivariant network has fewer redundant parameters that do not actually contribute to the network's prediction.
This can be further seen when we compare the Equivariant-ResNet-9 to a Equivariant-\gls{wrn}-40.
Even though both networks are equivariant, the larger \gls{wrn}-40 still has more unused parameters and subsequently is not able to improve on the smaller Equivariant-ResNet-9 (\textit{Eq.-ResNet-9}: $16k$ parameters; vs \textit{Eq.-\gls{wrn}-40}: $45k$ parameters).

\subsubsection{Sparser Gradient Vectors}\label{sec:sparse_grads}
\begin{figure}[t]
    \includegraphics[width=0.5\textwidth]{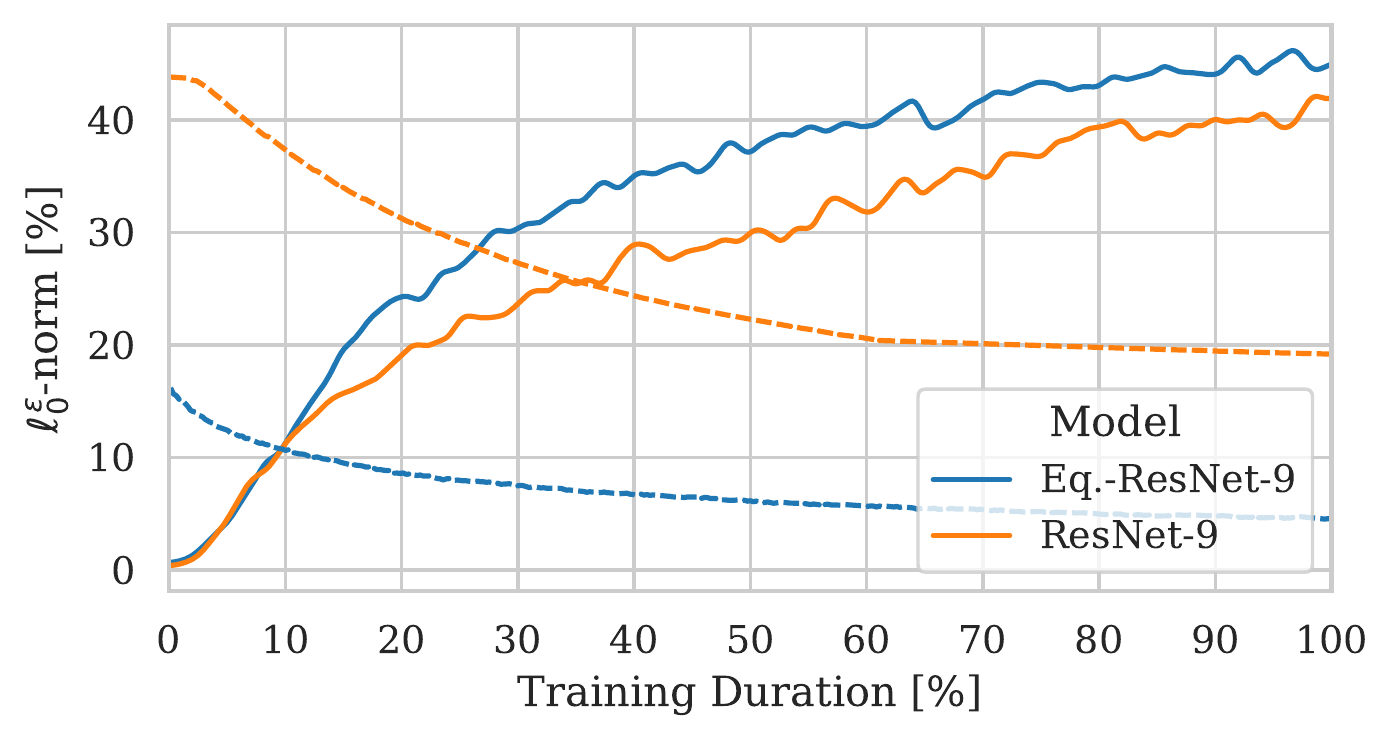}
\caption{The Equivariant-ResNet-9 converges to a better optimum during training, indicated by a higher gradient $\ell_{0}^{\epsilon}$ "norm" (\textit{continuous lines}), while also having a substantially lower percentage of parameters with values $< \epsilon$ (\textit{dashed lines}).}\label{fig:eval_sparsity}
\end{figure}

In addition to the network's weights, we also investigate how information is transmitted in the model.
If the features learned in the network show an increased representational efficiency, they should be able to learn useful information from the available data faster and thus require less time to converge to an optimum.
Analyzing the sparsity of the gradient vector in \cref{fig:eval_sparsity}, we see the \gls{ecnn}'s gradient converge faster during training (\ie more close-to-zero entries), updating fewer weights.
This can indicate that the model has incorporated most of the relevant information and has  approached a minimum. 
Moreover, gradient sparsity increasing more quickly during training also indicates a more efficient information transfer at an early stage.
This observation can not be made for the conventional ResNet-9, implying less efficient training.
The corresponding gradient sparsity increases slower and the model keeps updating more parameters throughout the training regime.
We consider additional research on the role of sparse gradient vectors in \gls{dp}, as \eg done in~\cite{zhu2021} and \cite{ito2022}, a promising direction for future work.
Our current results show that \gls{ecnn}s with increased designed sparsity have more parameters contributing to predictions and converge faster, with a smaller percentage of parameters to update during training.

\subsection{Improved Performance on Image Classification with Equivariant Convolutional Networks}\label{sec:results}
To validate that our notion of designed sparsity can in fact help to improve performance with \gls{dpsgd}, we compare the equivariant network to current \gls{sota} models on multiple common benchmark classification datasets when training from scratch under varying privacy parameters. The GPU hours ($h$) are measured on an NVIDIA A100 40GB.

\begin{table}[t]
\centering
\normalsize
\begin{tabular}{lllll}\toprule
& \multicolumn{4}{c}{CIFAR-10} \\ \cmidrule(r){2-5}
Model & $\varepsilon$ & Median & Std. Dev. & [$h$] \\ \specialrule{1pt}{1pt}{1pt}
\multirow{2}{*}{\textit{De} \etal (2022)} & 2 & 65.9 & \color{gray}(0.5) & $\ \ \ $-- \\
& 8 & 81.4 & \color{gray}(0.2) & $\ \ \ $-- \\
\multirow{2}{*}{\textit{De} \etal (\textit{reproduction})} & 2 & 62.6 & \color{gray}(0.62) & 41.3 \\
& 8 & 80.3 & \color{gray}(1.13) & 69.5 \\
\multirow{2}{*}{Equivariant-ResNet-9} & 2 & \textbf{71.8} & \color{gray}(0.71) & \textbf{5.8} \\
& 8 & \textbf{81.6} & \color{gray}(0.51) & \textbf{6.1} \\
\bottomrule \\
\toprule
& \multicolumn{4}{c}{CIFAR-100} \\ \cmidrule(r){2-5}
Model & $\varepsilon$ & Median & Std. Dev. & [$h$] \\ \specialrule{1pt}{1pt}{1pt}
\multirow{2}{*}{\textit{De} \etal (\textit{reproduction})} & 2 & 19.2 & \color{gray}(0.59) & 43.7 \\
& 8 & 40.8 & \color{gray}(0.15) & 76.1 \\
\multirow{2}{*}{Equivariant-ResNet-9} & 2 & \textbf{25.4} & \color{gray}(0.49) & \textbf{8.7} \\
& 8 & \textbf{48.4} & \color{gray}(0.30) & \textbf{8.9} \\
\bottomrule \\
\end{tabular}
\caption{CIFAR-10 and CIFAR-100 test accuracies of our Equivariant-ResNet-9 with symmetry group $D_4$ trained from scratch compared to the previous state of the art under $(\varepsilon,10^{-5})$-\gls{dp}. We report the median and standard deviation calculated across $5$ independent runs.}\label{tab:cifar}
\end{table}

\subsubsection{CIFAR-10} 
We begin with experiments on CIFAR-10, which is still considered a challenging dataset for training with \gls{dpsgd}.
The previous works' results were obtained by splitting the training set into $45k$ train and $5k$ validation samples.
Equivalently to these works, our stated test accuracy is achieved by training our model on the full training set and evaluating it once on the held-out test set.
Our equivariant models are benchmarked against the current \gls{sota} models on CIFAR-10 by \cite{de2022}.
We reproduce the previous \gls{sota} results with the exact same setup and code provided by the authors\footnote{\url{https://github.com/deepmind/jax_privacy}} on our hardware for a fair comparison of the prediction results and computation time.
The reproduced results have a difference in test accuracy of $\lessapprox 3\%$ to the results of the original paper, as similarly observed by \cite{sander2022}.
The exact hyperparameters and implementation details are summarized with further ablation studies in \cref{sec:params} and \cref{sec:appendix_results}.

\Cref{tab:cifar} shows, that our equivariant model with the $D_4$ group is able to substantially outperform the current \gls{sota} on CIFAR-10 by up to $9.2\%$ at $\varepsilon=2$.
Most notably, this result is achieved substantially faster with a decrease in computation time by $\approx 85\%$ ($35h$).
This is in large parts due to the reduced number of augmentations required.
Moreover, the corresponding model only consists of $256k$ parameters and is thus $35$ times smaller than the previous \gls{sota} model ($8.7M$ parameters less than the \gls{wrn}-40-4).
This superior performance of the Equivariant-ResNet-9 is consistent across all evaluated $\varepsilon$-values, and performs notably \textbf{even better under a tighter privacy budget} as seen in~\cref{tab:epsilon}.
This shows, that a sparser model architecture with a lower-dimensional gradient and thus less added noise is particularly beneficial for privacy-preserving applications.

\subsubsection{CIFAR-100}
The CIFAR-100 dataset is particularly interesting, as it has $10$ times fewer images per class compared to CIFAR-10.
This allows us to evaluate if the property of \gls{ecnn}s to learn better from less data also holds under \gls{dp}.
\Cref{tab:cifar} confirms, that this is indeed the case.
We train both models, the Equivariant-ResNet-9 and the \gls{wrn}-40-4 under the same setup as for CIFAR-10.
This time, however, our equivariant network is not only substantially better for lower $\varepsilon$-values, but also outperforms the \gls{wrn}-40-4 by more than $7\%$ under $\varepsilon=8$.
In addition, the computation performance remains superior, with a reduction in computation time from our equivariant network by $\approx 88\%$ ($67.2h$).
This performance can be attributed to the increased feature efficiency of \gls{ecnn}s (\textit{see \cref{sec:sparse_grads}}), which causes them to detect \textbf{more relevant input features} than non-equivariant models.
In addition to the benefits provided by their sparse model architecture (as seen for low $\varepsilon$-values for CIFAR-10), the feature efficiency of \gls{ecnn}s is a key characteristic making them useful for training with \gls{dpsgd} as privacy concerns mandate using as little data as possible.
To summarize, our \gls{ecnn}s achieve \gls{sota} performance on CIFAR-10 and CIFAR-100 with substantially smaller models and in a fraction of the computation time and are especially dominant under a tighter privacy budget.

\begin{table}[t]
\centering
\normalsize
\begin{tabular}{lllll}\toprule
& \multicolumn{4}{c}{Tiny-ImageNet-200} \\ \cmidrule(r){2-5}
Model & $\varepsilon$ & Median & Std. Dev. & [$h$] \\ \specialrule{1pt}{1pt}{1pt}
\textit{Klause} \etal (2022) & 10 & 19.4 & $\ \ \ $-- & $\ \ \ $-- \\
\gls{wrn}-40-4 & 8 & 27.8 & \color{gray}(0.55) & 696 \\
Equivariant-ResNet-9 & 8 & \textbf{34.1} & \color{gray}(0.08) & \textbf{156} \\
\bottomrule \\
\toprule
& \multicolumn{4}{c}{ImageNette} \\ \cmidrule(r){2-5}
Model & $\varepsilon$ & Median & Std. Dev. & [$h$] \\
\specialrule{1pt}{1pt}{1pt}
\textit{Klause} \etal (2022) & 9.88 & 67.1 & $\ \ \ $-- & $\ \ \ $-- \\
\gls{wrn}-40-4 & 8 & 70.0 & \color{gray}(1.79) & 45.7 \\
Equivariant-ResNet-9 & 8 & \textbf{75.6} & \color{gray}(2.20) & \textbf{19.7} \\
\bottomrule \\
\end{tabular}
\caption{Top-1 test accuracies on Tiny-ImageNet-200 and ImageNette for the Equivariant-ResNet-9 and a PyTorch reproduction of the \gls{wrn}-40-4 from \textit{De} \etal~\cite{de2022} compared to the previous \gls{sota} from \cite{klause2022} for $(\varepsilon, 8\cdot10^{-7})$-\gls{dp}.}\label{tab:imagenet}
\end{table}

\subsubsection{Tiny-ImageNet-200}
Large-scale image classification with \gls{dp} has recently come into focus, and promising initial results have lately been demonstrated on the ImageNet dataset~\cite{chrabaszcz2017, de2022, kurakin2022}. 
The key drawback of the aforementioned works is the fact that \gls{dp} training on ImageNet is \textit{exceedingly costly} in terms of computational budget required.
Efficient, yet accurate approaches are thus of great interest.
Even though our equivariant networks are not able to fully solve this issue, the previous experiments indicate that they offer a first step at achieving similar or better results to previous \gls{sota} approaches while reducing computation time.
The Tiny-ImageNet-200 dataset is particularly interesting, as it has fewer samples per class than the larger ImageNet-1k.
In addition to comparing to the previous \gls{sota} model~\cite{klause2022}, we also reproduce the approach from \textit{De} \etal with the \gls{wrn}-40-4 from~\cite{de2022}.
To compare, we use the substantially smaller Equivariant-ResNet-9 with the $D_4$ group and train both models for $12000$ update steps.
We also adapt the learning rate to $4$, due to the larger batch size.
For the experiment, we use the official validation set as the unseen test set.
\Cref{tab:imagenet} shows, that without further adaptations and hyperparameter tuning, the Equivariant-ResNet-9 achieves a Top-1 test accuracy of $34.1\%$, beating our baseline by $6.3\%$ and increasing the previous \gls{sota} by \cite{klause2022} by more than $14.7\%$ under a tighter privacy budget.
This is achieved while reducing the computation time by $\approx 77\%$ ($540h$).
We additionally also construct an equivariant version of the \gls{wrn}-40 model, the architecture previously used for \gls{sota} results~\cite{de2022}. 
As equivariant convolutions can be used in arbitrary network architectures, we only adapt the width of the equivariant model to $1$ instead of $4$, equivalent to the results in~\cref{sec:eval_sparsity} for the ResNet-9.
The model is also able to outperform the previous \gls{sota} result and the baseline model, but does not reach the performance of the smaller Equivariant-ResNet-9 (\textit{Eq.-\gls{wrn}-40}: $28.1\%$).
Looking at the $\ell_{0}^{\epsilon}$ "norm", we can see that both models use almost all of their parameters (\textit{Eq.-ResNet-9}: $2.5k$; \textit{Eq.-\gls{wrn}-40}: $7k$).
\textbf{Simply increasing the number of parameters in the equivariant models thus does not improve training proportionally.}
Instead, to boost performance with \gls{dpsgd}, future research could explore other designs that use their additional parameters efficiently.
These architectures with even greater designed sparsity may offer a promising approach for surpassing the performance of the proposed Equivariant-ResNet-9 model.

\subsubsection{ImageNette}
The preceding subsection demonstrated that \gls{ecnn}s, despite their compact model size, possess the capacity to learn complex features.
To assess their effectiveness as the image size is increased, we conduct further experiments on an additional ImageNet subset, ImageNette, featuring images with a dimension of $160$ by $160$ pixels.
As before, we compare the Equivariant-ResNet-9 to the previous \gls{sota} model by \cite{klause2022} and a custom baseline that reproduces the \gls{wrn}-40-4 setup from~\cite{de2022}.
In accordance with previous results, our equivariant network is able to outperform the previous \gls{sota} and our baseline in \cref{tab:imagenet}, by more than $5\%$.
In addition, the Equivariant-ResNet-9 takes less than half of the computation time compared to the baseline approach (reduction of $26h$).
Our proposed Equivariant-ResNet-9 thus outperforms all previous approaches independent of image size and number of images per class.
The accuracy improvement is particularly substantial under a tighter privacy budget, making the notion of designed sparsity a promising direction for further research on solving the privacy-utility trade-off.

\begin{table}[t]
\centering
\normalsize
\begin{tabular}{lrr}\toprule
& \multicolumn{2}{c}{Brier Score (\textit{lower is better})} \\ \cmidrule(l){2-3}
Dataset & ResNet-9 & Equivariant-ResNet-9 \\ \specialrule{1pt}{1pt}{1pt}
CIFAR-10 & 0.041 & \textbf{0.030} \\
CIFAR-100 & 0.0076 & \textbf{0.0069} \\
Tiny-ImageNet-200 & 0.004 & \textbf{0.0033} \\
ImageNette & 0.041 & \textbf{0.034} \\
\bottomrule \\
\end{tabular}
\caption{Compared to the standard ResNet-9, the Equivariant-ResNet-9 exhibits better model calibration, shown by a lower Brier score averaged across the test set under $(8,10^{-5})$-\gls{dp}.}\label{tab:brier}
\end{table}

\subsection{Privacy-Calibration Trade-Off}
While privacy guarantees are important in sensitive domains, evaluating other model characteristics should not be neglected in fields where trust in a model's predictions is essential.
In particular, it has been shown that \gls{dpsgd} has a negative effect on a model's uncertainty calibration~\cite{knolle2021, zhang2022}.
A model's overconfidence in its predictions when trained with \gls{dpsgd} can give a false sense of accuracy and prevent a reliable estimate of potential errors in the results.
We therefore exemplarily examine whether our proposed techniques are able to provide improved calibration in addition to higher accuracy by looking at the Brier score~\cite{brier1950}.
Indeed, on the CIFAR-10 test set, we found that the Equivariant-ResNet-9 had a $\approx 27\%$ lower Brier score compared to the non-equivariant counterpart (\cref{tab:brier}).
This result is consistent under the other evaluated datasets, with a reduction of on average $17\%$. 
The prediction improvements of the \gls{ecnn} thus do not further increase the overconfidence arising from \gls{dpsgd}.
Instead, the results indicate superior model calibration to standard convolutional models. 
While an in-depth investigation of the calibration of equivariant models is outside the scope of our current study, we consider this finding encouraging and intend to expand upon it in future work.

\section{Ablation Studies}
\subsection{Training Adaptations for Improved Performance of Sparse Models under DP}\label{sec:ablation}
To analyze the performance advantage of \gls{ecnn}s in more detail, we evaluate the impact of different hyperparameter choices in this section.
The results demonstrate key benefits of \gls{ecnn}s compared to conventional \gls{cnn} used for \gls{dp} training in previous works.

\begin{figure}[!t]
\centering
\includegraphics[width=.49\textwidth]{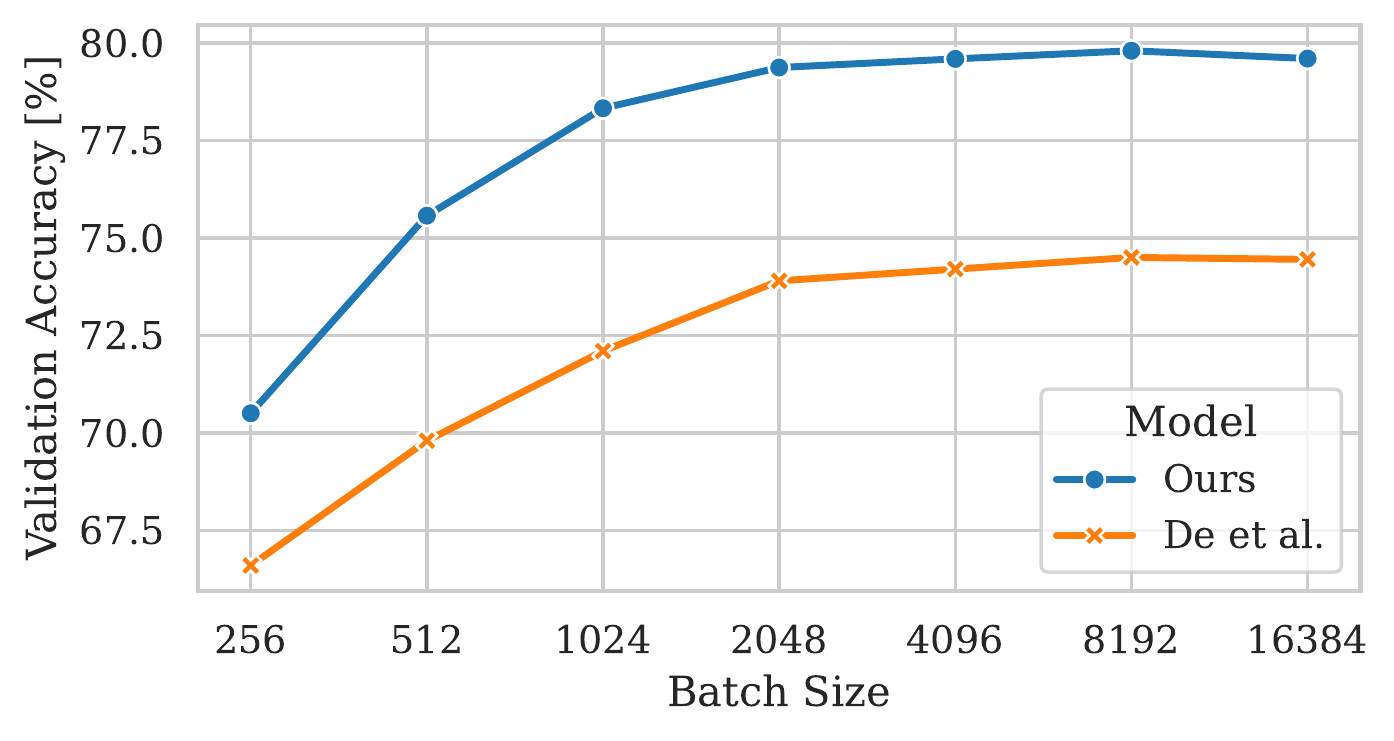}
\caption{The Equivariant-ResNet-9 benefits from increased batch sizes similar to the non-equivariant \gls{wrn}-16-4 from \textit{De}~\etal~\cite{de2022}, but has a substantially better validation accuracy across the board under $(8,10^{-5})$-\gls{dp}.}\label{fig:bs}
\end{figure}

\subsubsection{Improved Accuracy Across All Batch Sizes}
Previous work has generally shown that (very) large batch sizes lead to substantial improvements in accuracy for \gls{dp} training \cite{doermann2021, kurakin2022}.
We thus investigate whether the superior performance of \gls{ecnn}s is maintained across batch sizes or whether non-equivariant training is able to match our performance by batch size tuning.
We find that the latter is not the case.
In fact, our \gls{ecnn}s \textit{consistently outperformed the non-equivariant \gls{sota} models}, whereby larger batch sizes generally led to an increased accuracy, similar to the non-equivariant \gls{cnn}.
Interestingly, the accuracy gains through a batch size increase were \say{steeper} for \gls{ecnn}s compared to the baseline.
This is indicated by the slope of the curve between $256$ and $2048$ in \cref{fig:bs}, which is $\approx15\%$ steeper for the equivariant network.
We attribute this finding to the robustness of the features learned by equivariant kernels, which is synergistic with large batch sizes in making the gradient resilient to clipping and noise addition. 
In summary, the equivariant architecture outperformed the non-equivariant model across all batch sizes with an increase in validation accuracy of --on average-- $\approx5\%$.

\begin{figure}[t]
\centering
\includegraphics[width=.49\textwidth]{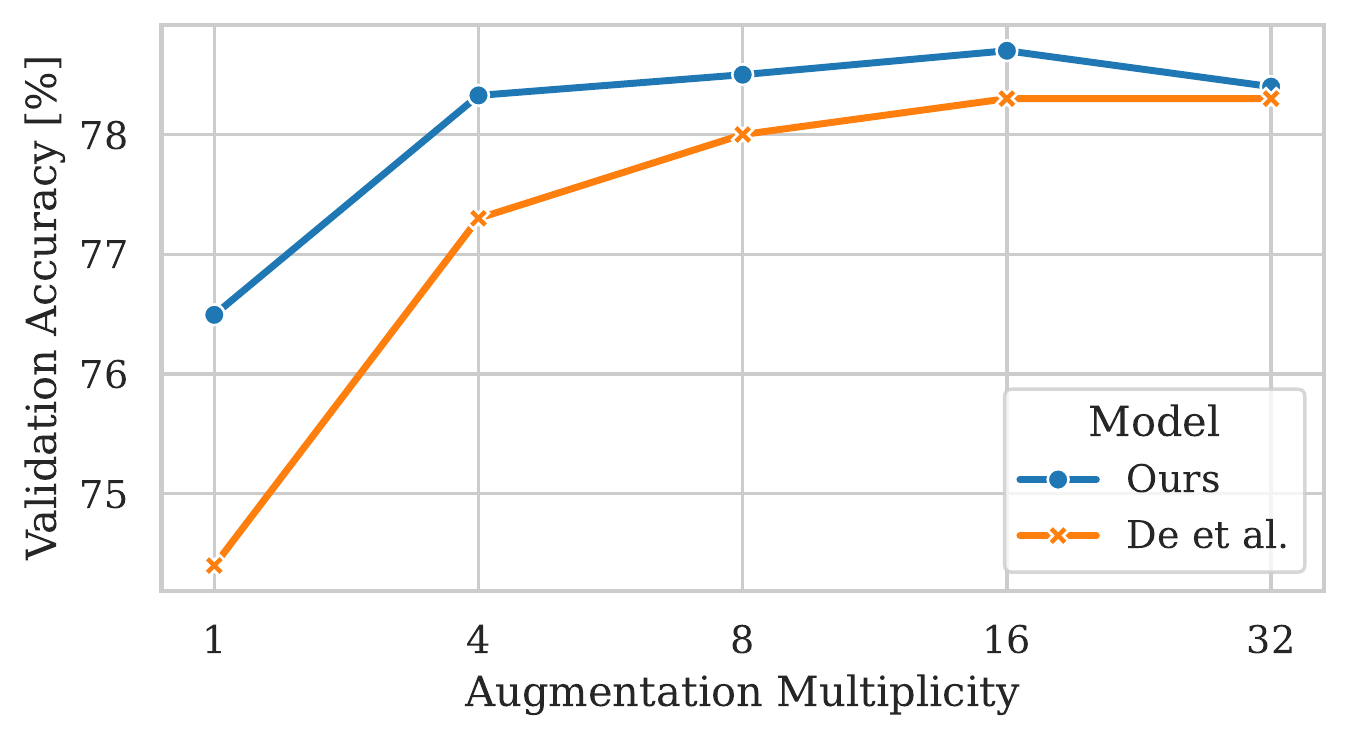}
\caption{The advantage of the equivariant network is maintained when adding augmentation multiplicities up to a value of $16$. Further increasing the number of augmentations only increases computation time but actually decreases performance.}\label{fig:aug_mult}
\end{figure}

\begin{table*}[!t]
\centering
\normalsize
\begin{tabular}{lllllll}\toprule
 &       & & \multicolumn{2}{l}{Test Accuracy [\%]} & &\\ \cmidrule{4-5}
 Model & Group & $\varepsilon$ & Median & Std. Dev. & Parameters & GPU Hours\\ \midrule
\textit{Dörmann} \etal (2021) & $\{e\}$ & 1.93 & 58.6 & \color{gray}(0.38) & $\ \ \ $--  & $\ \ \ $--   \\
\textit{De} \etal (2022) & $\{e\}$ & 2 & 65.9 & \color{gray}(0.5) & $8.9M$  & $\ \ \ $--  \\
\textit{De} \etal (\textit{reproduction}) & $\{e\}$ & 2 & 62.6 & \color{gray}(0.62) & $8.9M$  & 42.27  \\
\textit{Klause} \etal (2022) & $\{e\}$ & 2.89 & 65.6 & $\ \ \ $-- & $2.4M$  & $\ \ \ $-- \\
\textit{Tramèr and Boneh} (2021) & $\{e\}$ & 3 & 69.3 & \color{gray}(0.2) & $187k$  & $\ \ \ $-- \\
\midrule
\multirow{8}{*}{Equivariant-ResNet-9 (\textit{ours})}
& $C_{4}$ & 2 & 69.57 & \color{gray}(0.48) & $258k$ & 4.5\\ 
& $C_{8}$ & 2 & 68.97 & \color{gray}(0.19) & $256k$ & 5.8\\ 
& $C_{16}$ & 2 & 66.31 & \color{gray}(0.46) & $244k$ & 8.9\\ \cmidrule{2-7}
& $D_{4}$ & 2 & \textbf{71.86} & \color{gray}(0.71) & $256k$ & 5.8\\ 
& $D_{8}$ & 2 & 69.04 & \color{gray}(0.30) & $244k$ & 8.7\\ 
& $D_{16}$ & 2 & 67.68 & \color{gray}(0.39) & $238k$ & 14.8\\ \cmidrule{2-7}
& $SO(2)$ & 2 & 45.65 & \color{gray}(1.14) & $309k$ & 6.7\\ 
\bottomrule \\
\end{tabular}
\caption{CIFAR-10 test accuracy of our Equivariant-ResNet-9 with different symmetry groups trained from scratch compared to the previous state of the art without equivariance (\ie $\{e\}$). We report the median and the standard deviation calculated across $5$ independent runs. The GPU hours are measured on a NVIDIA A100 40GB. The highest accuracy is observed using the Equivariant-ResNet-9 and the $D_4$ group at only $256k$ parameters requiring only $5.8$ GPU hours of computation.}\label{tab:cifar_eps2}
\end{table*}

\subsubsection{Fewer Augmentations Required}
One of the main ingredients in the technique proposed by \cite{de2022} is the utilization of \textit{augmentation multiplicity}, \ie performing multiple augmentations per sample and averaging the resulting gradients before privatization. 
In this section, we address the question whether \gls{ecnn}s are able to supplant this technique (and thus drastically reduce the required computation time, which scales with the number of simultaneous augmentations).  
After all, equivariant convolutions are able to learn information \textit{decoupled from a feature's pose} and thus reduce the necessity of augmentations, especially of rotations and reflections, leading to an immediate reduction in the intense computational burden of augmentation multiplicity.
\cref{fig:aug_mult} shows, that the equivariant network still benefits from augmentations to some extent even though it does already incorporate pose information in its features.
This corroborates the finding by \textit{Weiler} \etal~\cite{weiler2019} in the non-\gls{dp} setting, who show that augmentations can further improve prediction performance of equivariant networks for dihedral groups.
Compared to the current state of the art on CIFAR-10, however, the equivariant models reach close-to-optimal performance with an augmentation multiplicity of only $4$, whereas the \gls{sota} model requires $8$ times as many simultaneous augmentations to achieve the same accuracy.
In fact, our results at $4$ augmentations \textit{outperform} the $32$ augmentation multiplicities of \cite{de2022} while drastically reducing the computation time by the same factor of $8$.
Corroborating the theoretical advantages of \gls{ecnn}s, diminishing returns set in earlier than for conventional \gls{cnn}.

\subsection{Hyperparameter Choices for Equivariant Layer}\label{sec:equ_hp}
As mentioned above, models utilizing the $D_4$ group exhibited the highest accuracy and were used for the experimental results above. 
To demonstrate how we arrived at this choice and how the chosen symmetry group influences the accuracy of the model, we evaluated our equivariant models with commonly used symmetry groups from literature \cite{weiler2019} under $(\varepsilon, \delta)=(2, 10^{-5})$-\gls{dp}.
As introduced in \cref{sec:background}, symmetry groups refer to mathematical groups that capture the symmetries or transformations that are applied on an input.
The symmetry group $C_8$ thus describes the group of rotations of $45$\textdegree on a planar space.
The continuous group $SO(2)$ extends this to all rotations in the 2-dimensional space.
We use irreducible representations and an angular frequency of $1$ to describe the $SO(2)$ group convolutions.
For an in-depth analysis of other frequency groups of the $SO(2)$ group, we refer to~\cref{sec:o2_frequencies}.
Based on the results from~\cref{sec:ablation}, we reduce the model layout from $(64, 128, 256)$ channels to $(16, 32, 64)$.
To maintain comparability, the number of channels per layer and thus total number of parameters is fixed according to our approach described in \cref{sec:method}.
The results in \cref{tab:cifar_eps2} show that we achieve our best median test accuracy on CIFAR-10 with $71.86\%$ for the dihedral group $D_4$.
The dihedral groups perform (on average) slightly better than the cyclic groups, probably due to the intrinsic horizontal symmetry of the images in the dataset which are better captured by its capability to represent reflections.
This is corroborated by the fact that increasing the rotation order $N$ did not improve prediction performance.
As the network uses kernels of size $3\times3$, this phenomenon could also be attributed to the restrictions in discretizing small rotations on this kernel size without losing information. 
We observe a similar effect with the continuous rotations in $SO(2)$, which perform substantially worse than all other groups.
This is in line with the non-private experiments conducted by~\cite{weiler2019}, indicating that the kernel constraint is too restrictive and the lack of expressiveness cannot be compensated by the more pronounced equivariant properties of the kernel.
We consider combining our technique with larger receptive fields, \eg through \textit{atrous} (dilated) convolutions, a promising future work direction.
Notably, all discrete groups $C_N$ and $D_N$ were able to outperform previous \gls{sota} models.
The $C_4$ and $D_4$ groups in particular showed the best results, offering the best trade-off between accuracy and computation time out of all candidates.
We thus recommend favoring these groups over continuous groups in practice.

\section{Conclusion}\label{sec:conclusion}
The broad application of private deep learning has --so far-- been impeded by privacy-utility trade-offs. 
Recent works have partially addressed these limitations and presented techniques to bridge the accuracy gap but introduced a new trade-off between accuracy and efficiency.
Ultimately, we contend that \textit{both} trade-offs must be addressed to facilitate large-scale research in \gls{dp} deep learning.
The remarkable performance gains that Equivariant \gls{cnn} enable are an important step towards this goal.
Their capability to outperform previous approaches in a low-data regime and under a tight privacy budget, as well as their improved calibration and capability to capture intrinsic image symmetries, renders them particularly interesting for \gls{dpsgd}.

With extensive benchmark experiments, we showed that sparse model designs are promising to also overcome computational overhead concerns in \gls{dp}.
In a time such as the present, where breakthroughs are usually achieved through solving engineering problems on large-scale systems, targeting ways to render \textit{the foundations of deep learning itself more efficient} is in our opinion a promising and sustainable direction for solving long-term challenges.
The introduction of additional structural prior information, such as the presence of symmetries in images, tackles exactly such challenges.
In addition, we regard the provision of formal guarantees of model behavior, which both equivariance and \gls{dp} represent, as a solid foundation for systems that fulfill the notion of \say{trustworthy AI}.\\

%%%%%%%%% REFERENCES
\onecolumn
\twocolumn
\small
\bibliographystyle{IEEEtran}
\bibliography{arxiv}

%%%%%%%%% APPENDIX
\onecolumn
\twocolumn
\normalsize
\appendix
\section{Appendix}
\subsection{Implementation Details} \label{sec:params}
All experiments in this paper except for the reproduction of~\cite{de2022} are implemented in \texttt{PyTorch}.
The privacy accounting is done with \texttt{Opacus} and additional performance improvements are achieved through vectorization with \texttt{func\-torch}.
The aforementioned reproduction is implemented with the code taken from the official repository of the original authors in \texttt{JAX/Haiku}.
The hyperparameters used for our equivariant networks and the reproduction are summarized in~\cref{tab:hyperparams}.
The equivariant models are trained for $2160$ update steps with \gls{dpsgd} and a clipping norm and learning rate of $2.0$, an exponential moving average decay of $0.999$, a batch size of $8192$ and $4$ augmentation multiplicities  (\textit{original image + 3 augmentations}). 
We use random reflections and cropping with a two-sided reflection padding of $4$ pixels for our augmentations.

\begin{table}[h]
\centering
\begin{tabular}{lcc}\toprule
 &  CIFAR & Tiny-ImageNet-200 \\ \midrule
Optimizer & SGD & SGD \\
Aug. Multiplicity & 4 & 4  \\ 
Batch Size    & 8192 & 16384  \\
Clipping Norm & 2.0 & 2.0  \\ 
Learning Rate & 2 & 4  \\ 
Noise Multiplier $\sigma$ & 5.0 & 12.5  \\ 
Number Updates  & 2160 & 12000  \\ 
\bottomrule \\
\end{tabular}
\caption{Hyperparameters used for measuring the accuracy when training from scratch in all of our equivariant experiments on test sets.}
\label{tab:hyperparams}
\end{table}

\subsection{Additional Experimental Results}\label{sec:appendix_results}
\subsubsection{Results Across Different $\varepsilon$-Values}
For easier reproducibility, the exact results for different $\varepsilon$-values are given in \cref{tab:epsilon}.
All experiments are run with the setting described in \cref{sec:results} and hyperparameters summarized in~\cref{tab:hyperparams}.
The standard deviation for our equivariant network is measured across $5$ independent runs, while the reproduction was run $3$ times.

\begin{table}[h!]
\centering
\begin{tabular}{llll}
\toprule
 &  & \multicolumn{2}{l}{Test Accuracy [\%]}\\ \cmidrule{3-4}
 Model & $\varepsilon$ & Median & Std. Dev. \\ \midrule
\multirow{6}{*}{Equivariant-ResNet-9 (\textit{ours})} & 1 & 60.59 & \color{gray}(1.40) \\ 
& 2 & 72.10 & \color{gray}(0.68) \\ 
& 3 & 75.96 & \color{gray}(0.38) \\ 
& 4 & 78.27 & \color{gray}(0.15) \\ 
& 6 & 80.26 & \color{gray}(0.26) \\ 
& 8 & 81.62 & \color{gray}(0.51) \\ 
\midrule
\multirow{5}{*}{\gls{wrn}-40-4  (\textit{reproduction})} & 1 & 52.80 & \color{gray}(0.45) \\ 
& 2 & 62.63 & \color{gray}(0.62) \\ 
& 4 & 72.42 & \color{gray}(0.84) \\ 
& 6 & 77.78 & \color{gray}(0.45) \\ 
& 8 & 80.38 & \color{gray}(1.13) \\ 
\bottomrule \\
\end{tabular}
\caption{CIFAR-10 median test accuracies under different $(\varepsilon, 10^{-5})$ values measured across $5$ independent runs for the Equivariant-ResNet-9 with symmetry group $D_4$ and $3$ independent runs for the \gls{wrn}-40-4 from \cite{de2022}.}
\label{tab:epsilon}
\end{table}

\subsubsection{Frequencies of the $SO(2)$ Group}\label{sec:o2_frequencies}
For the continuous $SO(2)$ group, instead of the number of discrete rotations, we have to choose the angular frequency used to construct the kernel as an additional hyperparameter.
The choice not only changes the expressiveness of the kernel but also impacts the number of parameters that scale consistently with the angular frequency.
As in earlier sections, we thus adapt the model width to have a comparable number of parameters.
\cref{tab:angular_frequencies} shows, that when training under \gls{dp}, the increased expressiveness does slightly increase prediction performance, with a frequency of $3$ performing best under $(2, 10^{-5})$-\gls{dp}.
However, conforming with the results in \cref{sec:equ_hp}, the $SO(2)$ group performs worse than the discrete groups independent of the chosen angular frequency.
Our results thus suggest to use the discrete $C_N$ and $D_N$ groups when training \gls{ecnn}s with \gls{dpsgd}.

\begin{table}[h!]
\centering
\begin{tabular}{llll}\toprule
$\varepsilon$ & Frequency & Parameters & Test Accuracy [\%] \\ \midrule
\multirow{3}{*}{$2$} & $1$ & $243k$ & 43.69  \\ 
& $2$ & $251k$ & 44.65  \\ 
& $3$ & $309k$ & \textbf{45.65}  \\ \cmidrule{1-4} 
 \multirow{3}{*}{$8$} & $1$ & $243k$ & 53.61  \\ 
& $2$ & $251k$ & \textbf{55.12}  \\ 
& $3$ & $309k$ & 54.23  \\ 
\bottomrule \\
\end{tabular}
\caption{CIFAR-10 test accuracies for different angular frequencies of the steerable kernel for the $SO(2)$ group for an adapted layout with similar number of parameters.}
\label{tab:angular_frequencies}
\end{table}

\end{document}